\documentclass[runningheads]{llncs}
\usepackage[T1]{fontenc}
\usepackage{graphicx}
\usepackage{cite}
\usepackage{amsmath,amssymb,amsfonts}
\usepackage{algorithm}
\usepackage{algorithmic}
\usepackage{comment}
\usepackage{textcomp}
\usepackage{subfigure}
\usepackage{hyperref}
\usepackage{xcolor}
\usepackage{float}
\usepackage{multirow}
\usepackage{bbm}

\begin{document}
\title{OpenPath: Open-Set Active Learning for Pathology Image Classification via 
Pre-trained Vision-Language Models}
%

\titlerunning{OpenPath}
\author{Lanfeng Zhong\inst{1,2}  \and Xin Liao\inst{3} \and Shichuan Zhang\inst{4} \and Shaoting Zhang\inst{1,2} \and Guotai Wang\inst{1,2}}

\authorrunning{L.Zhong et al.}


\institute{University of Electronic Science and Technology of China, Chengdu, China \and Shanghai Artificial Intelligence Laboratory, Shanghai, China \and Department of Pathology, West China Second University Hospital, Sichuan University, Chengdu, China \and Department of Radiation Oncology, Sichuan Cancer Hospital and Institute, University of Electronic Science and 
Technology of China, Chengdu, China \email{guotai.wang@uestc.edu.cn}}
\maketitle              
\begin{abstract}
Pathology image classification plays a crucial role in accurate medical diagnosis and treatment planning. Training high-performance models for this task typically requires large-scale annotated datasets, which are both expensive and time-consuming to acquire. Active Learning (AL) offers a solution by iteratively selecting the most informative samples for annotation, thereby reducing the labeling effort. However, most AL methods are designed under the assumption of a closed-set scenario, where all the unannotated images belong to target classes. In real-world clinical environments, the unlabeled pool often contains a substantial amount of 
Out-Of-Distribution (OOD) data, leading to low efficiency of annotation in traditional AL methods. Furthermore, most existing AL methods start with random selection in the first query round, leading to a significant waste of labeling costs in open-set scenarios.
To address these challenges, we propose OpenPath, a novel 
open-set active learning approach for pathological image classification leveraging a 
pre-trained Vision-Language Model (VLM). In the first query, we propose 
task-specific prompts that combine target and relevant non-target class prompts to effectively select In-Distribution (ID) and informative samples from the unlabeled pool. In subsequent queries, Diverse Informative ID Sampling (DIS) that includes 
Prototype-based ID candidate Selection (PIS) and Entropy-Guided Stochastic Sampling (EGSS) is proposed to ensure both purity and 
informativeness in a query, avoiding the selection of OOD samples. 
Experiments on two public pathology image datasets show that OpenPath significantly enhances the model's performance due to its high purity of selected samples, and outperforms 
several state-of-the-art open-set AL methods.
The code is available at \href{https://github.com/HiLab-git/OpenPath}{https://github.com/HiLab-git/OpenPath}.

\end{abstract}

\section{Introduction}
Pathology images provide crucial information for cancer detection, severity assessment, and treatment planning~\cite{schiffman2015early}. 
Recent advances in deep learning, particularly with Convolutional Neural Networks (CNNs)~\cite{resnet} and Vision Transformers (ViTs)~\cite{vit2021,swin-vit}, have improved 
pathology image classification accuracy and efficiency. 
However, these models depend 
on large annotated datasets, which are costly and time-consuming to create, often 
requiring expert pathologist input.


Active Learning (AL) aims to minimize annotation costs by selecting the most 
informative samples. Traditional AL methods based on 
uncertainty~\cite{schmidt2024focused-AL,entropy_al,zhong2025unisal}, 
diversity~\cite{arthur2007kmeans++,coreset}, and hybrid 
strategies~\cite{wang2023active} perform well in closed-set scenarios, where 
unlabeled data contains only target classes. However, in real-world applications, 
unlabeled data often includes both In-Distribution (ID) and Out-of-Distribution 
(OOD) classes, leading to poor performance of traditional methods, as they may select numerous OOD samples that contribute little to model improvement.
In pathological image analysis, high-resolution Whole Slide Images (WSIs) are typically divided into smaller patches representing various tissue types, and annotating non-target (i.e., OOD) classes is often unnecessary for diagnosis. For example, in colorectal cancer diagnosis, 
non-target classes like muscle tissue are irrelevant to cancerous regions and do not necessarily require annotations.
This scenario, where the unlabeled pool contains both target and 
non-target samples, is known as an ``open-set AL problem'',
aiming at maximizing the purity, i.e., the ratio of ID images in the query
to avoid the annotation of OOD images.

Recently, several open-set AL methods have been proposed for natural and pathology images. For 
example, Ning et al.~\cite{LfOSA} 
developed a classifier with 
$C$ ID classes and an explicit OOD class, prioritizing samples with high ID-class activation scores to minimize OOD selection. 
OpenAL~\cite{OpenAL} 
uses a two-stage method that selects ID samples 
by feature distance and uncertainty to select ID samples with high uncertainty.
However, these methods neglect to consider both diversity and uncertainty simultaneously. Moreover, these methods
do not effectively address the cold start problem, as they rely on 
random selection 
for the first query, leading to a significant waste of annotation budget by 
including many OOD samples that are irrelevant to the target task.

In this paper, we introduce \textbf{OpenPath}, a novel and efficient open-set active 
learning method for pathology image classification. The contribution is two-fold. 
First, we propose a Vision-Language 
Model (VLM)-based cold start method to effectively query ID images in the first 
round. To achieve this, we leverage GPT-4 to suggest task-specific OOD classes to represent non-target classes that may present in the unlabeled pool. These ID and potential OOD classes enable prompting the VLM to perform zero-shot inference to obtain 
pseudo-labels for unlabeled images, which allows for retaining
ID class candidates.
Second, we introduce a Diverse Informative ID Sampling (DIS) 
for subsequent queries when the model is trained.
It starts with Prototype-based ID candidate Selection (PIS) to distinguish 
ID from OOD samples, and then uses
Entropy-Guided Stochastic Sampling (EGSS) to ensure
that selected samples are both diverse and informative by considering random 
batches and entropy. 

We performed extensive experiments on the public
CRC100K~\cite{kather2016multi} and SkinTissue datasets~\cite{skintissue}. The results showed that OpenPath significantly 
enhances the purity of queries and achieves superior performance 
with the same annotation budget, outperforming current state-of-the-art open-set AL methods.

\begin{figure}[t] \centering
\includegraphics[width=\textwidth]{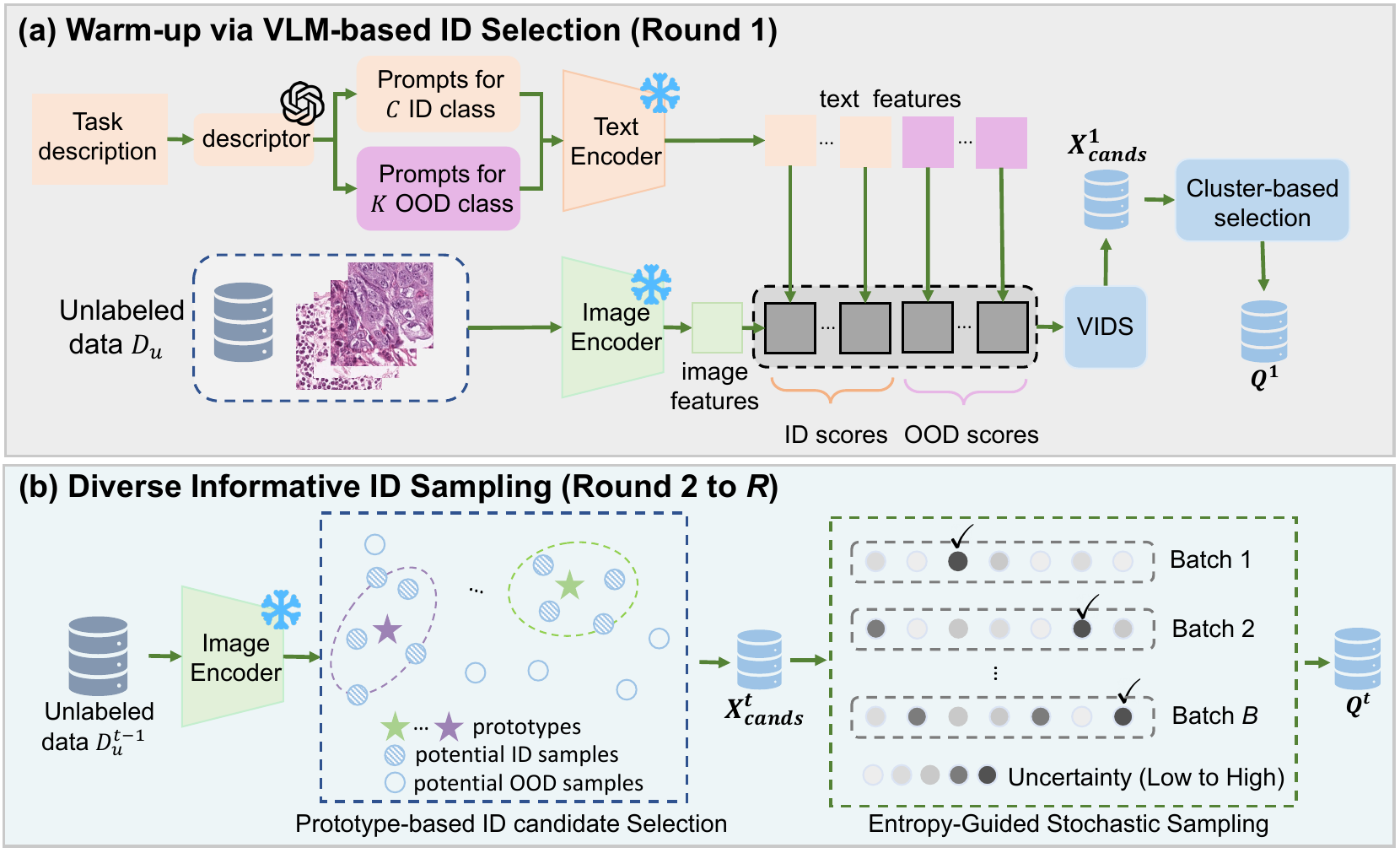}
\caption{Overall framework of our OpenPath.
VIDS: VLM-based ID Sample Selection, which ensures a high ID ratio in the first query. $Q^{t}$: samples selected in the t-th query.
}
\label{fig:overall_2}
\end{figure}

\section{Methods}

\subsection{Problem Formulation and Method Overview}
The open-set AL problem addressed in this study focuses on selecting samples to annotate for training a classification model $f$ for $C$ target classes, and the unlabeled pool contains both samples from the $C$ target (ID) classes and some non-target (OOD) classes. 
We consider a large-scale unlabeled pathology image dataset with $N$ samples \( \mathcal{D}_u = \{ x_i \}_{i=1}^N \). After the $t$-th query ($t \ge 1$),  we have a labeled set \( \mathcal{D}_l^t = \{ (x_i, y_i) \}_{i=1}^{N_{l}^t} \), where $N^t_l$ is the size of $D^t_l$.
We define the annotation budget per round as \( L \) and the number of 
query rounds as \( R \).
The $t$-th query set is denoted as $Q^t$, and its subset for ID and OOD are denoted as $\hat{Q}^t$ and $\tilde{Q}^t$, where $\hat{Q}^t$ $\cap$ $\tilde{Q}^t$ = $\emptyset$, and $\hat{Q}^t \cup \tilde{Q}^t = Q^t$. The accumulated labeled set after the $t$-th query is $D^t_l = Q^1 \cup Q^2 \cup ... \cup Q^t$.
During annotation, experts provide fine-grained labels for ID samples, 
while OOD samples are labeled as “non-target” without specifying a particular class.

As shown in Fig.~\ref{fig:overall_2}, to maximize the selection of ID samples, our method begins with a warm-up phase, where task-specific prompts are generated by GPT to guide a pre-trained vision-language model in identifying ID and OOD samples. Samples with high ID scores are retained for further selection. In subsequent query rounds, the ID candidates are selected by distance from a sample to the prototypes of ID classes in the feature space. 
To further reduce redundancy and improve the informativeness of ID candidates, 
we randomly split them into several batches and select the most uncertain samples in each batch.

\subsection{Warm-up via VLM-based ID Sample Selection}

Most existing AL methods~\cite{OpenAL, LfOSA, entropy_al} 
rely on random selection for the initial query. 
However, when the
unlabeled pool includes both ID and OOD data, these methods can unintentionally
select a large number of OOD samples. To address this, we propose a 
VLM-based ID Sample Selection (VIDS) to prioritize ID samples and ensure a high ID ratio (i.e., purity) in the first query.

\subsubsection{VLMs with Task-Specific OOD Prompts}

VLMs like CLIP~\cite{clip} have zero-shot inference ability by computing the similarity between an image feature and different text prompts. Let $E_{img}$ and $E_{text}$ denote the encoders for image and texts respectively, with $D$ denoting the feature dimension.
For an input image $x$, its
image feature representation is \( z = E_{img}(x) \). 
Let $T_c$ denote the name of the $c$-th class, the text prompt for that class is given by a specific template, such 
as \( \mathcal{T}_c = \) ``\texttt{An H\&E image of \{T$_c$\}}''. 
The text feature \( g_c \in \mathbb{R}^D 
\) is \( g_c = E_{text}(\mathcal{T}_c) \).
While VLMs require knowledge of all class names for inference, in the open-set AL scenario, we only know names of ID classes, and OOD class names vary across applications, making manual defining them impractical. Thus,
automatically suggesting potential OOD classes for a specific dataset is desirable.



To achieve this, we create task-specific text prompts that include task-specific information, helping the VLM effectively differentiate between ID and OOD data. 
Specifically, GPT-4 is used to generate these OOD classes because it can incorporate precise domain knowledge into textual descriptions, enhancing the VLM's ability to guide the sample selection. 
We design a question template for GPT-4 as {\texttt{``The task is \{TASK\}, focusing 
on target classes including \{ID classes\}. Please provide K
other tissue categories that may be \\ present in this task''}. 
By filling in the specific task and ID classes, relevant OOD classes can be obtained.
The total number of classes for the VLM is \( C' = C + K \), where each class name is converted into a prompt \( \mathcal{T}_c \) for the \( c \)-th class (\( c = 0, 1, \dots, C+K-1 \)) using the template mentioned above.


\subsubsection{Representative ID Samples Selection}
With the designed task-specific OOD prompts, the class probability \( p_c \) for each image \( x \) is computed as:
\begin{equation}
    p_c = \frac{e^{sim(z, g_c)/\tau}}{\sum_c e^{sim(z, g_c)/\tau}}
    \label{eq1}
\end{equation}
where $sim$ and $\tau$ denote the cosine similarity and temperature coefficient, respectively.
Then the 
pseudo-label \(\hat{y}\) is obtained via argmax over the probability vector \( \mathbf{p}=[p_0, p_1, ..., p_{C+K-1}] \). 
Samples with pseudo-labels corresponding to ID classes are selected as ID candidates for the first query round:
\begin{equation}
    X_{cands}^1 = \{x_i \mid x_i \in \mathcal{D}_{u},~~\hat{y}_i < C \}
    \label{eq_init_1}
\end{equation}

As the size of $X^1_{cands}$ is much larger than $L$, we then select the representative ones for labeling. 
KMeans++~\cite{arthur2007kmeans++} is employed to cluster samples in $X^{1}_{cands}$. 
With the number of clusters set to the budget size \( L \), the nearest sample to each centroid is selected:
\begin{equation}
    Q^{1} = \{ \underset{x_i \in X_{cands}^1}{\operatorname{argmin}} ||z_i-O_j||; ~j=1, 2, ..., L\}
    \label{eq_init_2}
\end{equation}
where \( O_j \) is the \( j \)-th clustering centroid, and \( ||\cdot|| \) denotes the L2 distance. The labeled pool is updated as \( \mathcal{D}_l^1 = Q^1 \), and the unlabeled pool is $\mathcal{D}_u^1=\mathcal{D}_u-\mathcal{D}_l^1$.

\subsection{Diverse Informative ID Sampling in Subsequent Queries}

\subsubsection{Prototype-based ID Candidate Selection}
In the subsequent query rounds $(t \ge 2)$, as the classifier $f$ has been trained with previously queried samples in $D^{t-1}_l$, it provides a more adaptive feature representation of the target dataset than the VLM, therefore 
we update the prototypes of ID classes based on the trained classifier. 
The prototype of each ID class \(c \in \{0, \ldots, C-1\}\) is calculated as the average feature \(\bar{z}^c\) for that class based on $\mathcal{D}_l^{t-1}$. 
Inspired by OOD detection~\cite{simple_ood} where the distance to ID samples can be used to calculate OOD scores, we  measure the proximity of unlabeled samples to ID 
classes for selecting ID candidates.
Let $d_i$ denote the distance of $x_i$ to its nearest ID prototype: $d_i = \min_{c \in \{0, 1, \dots, C-1\}} \left(1 - sim(z_i, \bar{z}^c)\right)$.
A low $d_i$ value indicates that the sample is more likely to belong to ID classes. 
Let \( d_M \) denote the \( M \)-th percentile of \( d_i \) across $D^{t-1}_u$. We use \( d_M \) as a threshold to derive ID candidates set $X_{cands}^t$ for the $t$-th ($t\ge2$) round from \( \mathcal{D}^{t-1}_u \), where the size of $X_{cands}^t$ is $ M\% \cdot|\mathcal{D}^{t-1}_u|$:
\begin{equation}
    X_{cands}^t = \{x_i~|~x_i \in 
    \mathcal{D}_u^{t-1}, ~d_i \leq d_M\}
    \label{eq_loop_1}
\end{equation}

\subsubsection{Entropy-Guided Stochastic Sampling}
Since $|X^t_{cands}|$ is larger than $L$, 
we need to further select $L$ samples from \( X^t_{{cands}} \) that are not only highly informative but also diverse, as both properties are crucial for AL~\cite{al_survey}.
Existing AL methods~\cite{entropy_al,OpenAL} mainly select the
most uncertain samples based on the entropy \( \mathcal{H}(x_i) = -\sum_{c=0}^{C-1} p^c_i \log(p_i^c) \), where $p_i^c$ is the probability of class $c$ for $x_i$. However, only selecting uncertain samples overlooks the diversity of queried samples, and easily leads to redundancy in the query.
To address this issue, we propose Entropy-Guided Stochastic Sampling (EGSS) to encourage informativeness and diversity simultaneously. 
Inspired by~\cite{stochastic_b}, which leverages random batching to ensure diversity, we refine this approach by selecting only the top uncertain samples within each batch rather than choosing entire batches with the highest mean entropy. 
As shown in Fig.~\ref{fig:overall_2}(b), 
we divide the samples in $X_{cands}^t$ into 
$B$ batches. From the $b$-th batch, we select the top 
$I$ samples with the highest uncertainty to form $Q^t_b$, where 
$I=L/B$. This results in \( L \) selected samples for annotation $Q^t = Q^t_1 \cup Q^t_2 \cup \ldots \cup Q^t_B$.

\section{Experiments}
\subsection{Implementation Details}
\noindent
\textbf{Datasets} 
We evaluated the effectiveness of OpenPath on the 
CRC100K~\cite{kather2016multi} and SkinTissue~\cite{skintissue} datasets.
CRC100K is a public colorectal cancer pathology dataset containing 9 classes: adipose (ADI), background (BACK), debris (DEB), lymphocytes (LYM), mucus (MUC), smooth muscle (MUS), normal colon mucosa (NORM), cancer-associated stroma (STR), and adenocarcinoma epithelium (TUM). For clinical relevance, we designated LYM, NORM, and TUM as ID classes for the open-set AL setting, treating the remaining classes as OOD. The dataset comprises 100,000 training patches (224×224 pixels at 0.5 microns per pixel) extracted from 86 tissue slides, with classes balanced at 9\%–12\% each.
Following~\cite{OpenAL}, 2608 ID class samples in the testing set 
were used for evaluation.

SkinTissue~\cite{skintissue} is a skin cancer pathology image dataset comprising 16 classes with a high imbalance. The classes and their ratios are: necrosis (1.84\%), 
naevus (8.90\%), skeletal muscle (6.92\%), squamous cell carcinoma (7.63\%), sweat glands 
(2.84\%), vessels (1.20\%), elastosis (0.15\%), chondral tissue (4.99\%), hair follicle (1.61\%), 
epidermis (11.71\%), nerves (1.34\%), subcutis (8.28\%), melanoma (8.74\%), epithelial basal cell 
carcinoma (BCC, 7.77\%), dermis (17.84\%), and sebaceous glands (8.16\%).
Melanoma and BCC were designated as ID classes (16.51\%), with all other classes treated as OOD.
It comprises 88,171 image patches for training, manually extracted from 1,166 patients. For testing, we used 3619 ID class samples in the official testing set. 
Each image patch covers an area of 100×100\(\mu\text{m}\)
(395×395 pixels).

\noindent
\textbf{Implementation Details} 
In the experiments, we used
BioMedCLIP~\cite{biomedclip}, which employs a 
ViT-B/16~\cite{vit2021} image encoder with an output dimension of 768. The 
classifier $f$ consisted of BioMedCLIP’s image encoder and a two-layer fully 
connected classification head (256 and $C$ output nodes).
The loss function was cross-entropy loss.
In each query round, the model was trained using cross-entropy loss and the SGD optimizer (with weight decay set to \( 8 \times 10^{-4} \)), for 30 epochs on CRC100K and 50 epochs on SkinTissue, with all model parameters being optimized during training.
The learning rate started at \( 10^{-3} \) and decayed by a factor of 0.1 every 20 epochs. 
$B$ was set to 10 following~\cite{stochastic_b}. For CRC100K and SkinTissue, 
$M$ was set to 25 and 7, $L$ was 50 and 200
respectively based on dataset-specific ID/OOD ratios and task complexities. 
Query round $R$ was set to 5.
We repeated the experiments five times with different random seeds and reported the average results for all methods.

\noindent
\textbf{Evaluation Metrics}
To evaluate model performance, we used Model Accuracy (MAcc) on the testing set. 
Query Precision (QP) was employed to measure the purity of the queried batch, 
defined as the ratio of ID samples in the query. Additionally, Accumulated 
Query Recall (AQR) quantifies the ability to retrieve ID samples across the entire dataset after the $t$-th query.
\begin{equation}
    \text{QP}_{t} = \frac{|\hat{Q}^{t}|}{|\hat{Q}^{t}|+
    |\tilde{Q}^{t}|},~\text{AQR}_{t} = \frac{\sum_{j=1}^{t}|\hat{Q}^{j}|}{|\hat{\mathcal{D}}_{u}|}
    \label{eq:precision}
\end{equation}
where $|\hat{\mathcal{D}}_{u}|$ is the number of real ID samples in the training set.

\begin{figure}[t] \centering
\includegraphics[width=\textwidth]{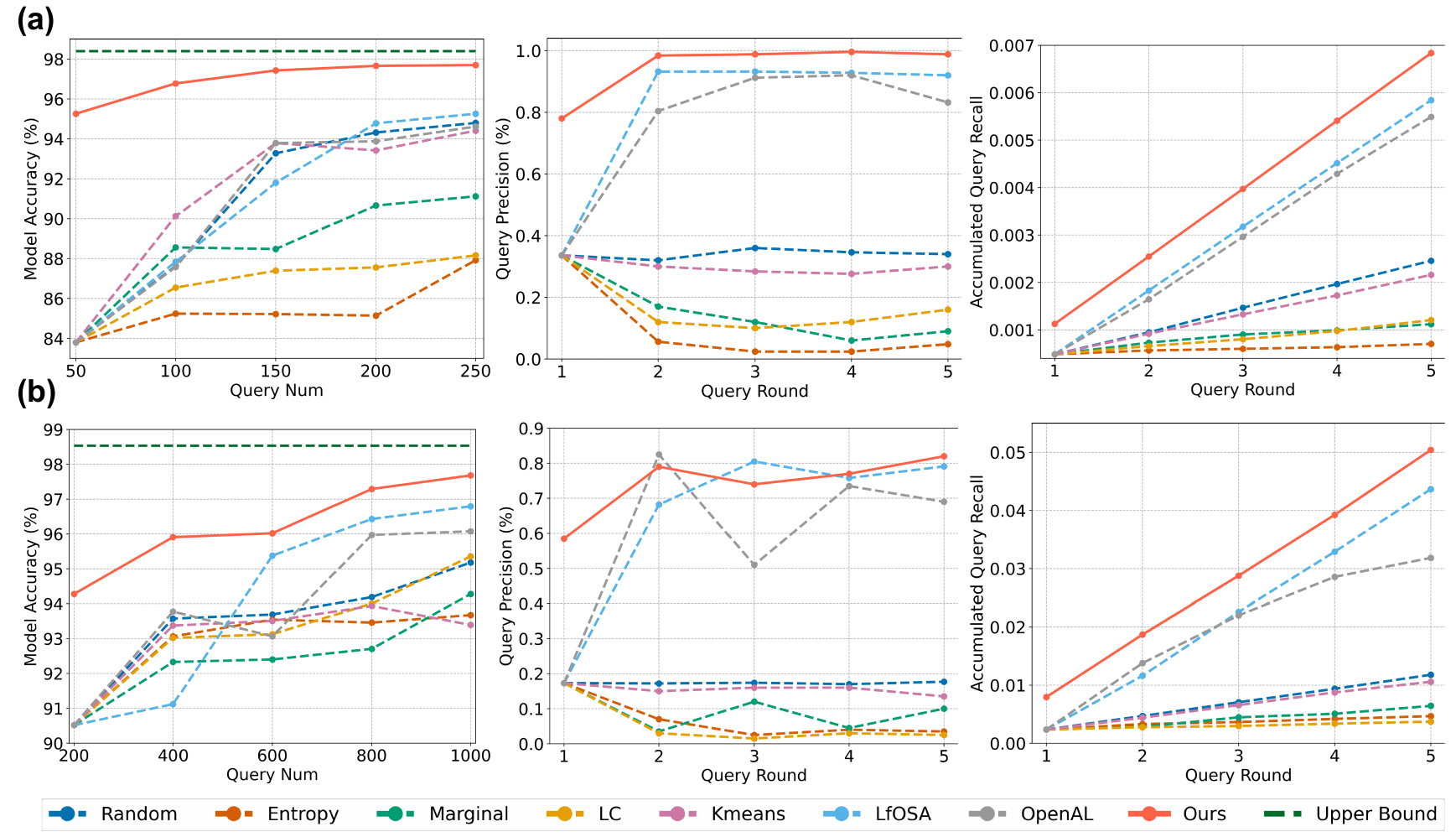}
\caption{{Comparison between OpenPath and existing methods in terms of model accuracy (left), query precision (middle), and accumulated query recall (right). 
(a) CRC100K dataset.
(b) SkinTissue dataset. Upper bound denotes training $f$ using $\hat{\mathcal{D}}_u$.}}
\label{fig:comparision_crc100k_36}
\end{figure}

\subsection{Comparisons with State-of-the-art Methods}
For comparison, we established a baseline based on random sample 
selection for each query round. Our OpenPath was 
compared with six existing AL methods:
1) Kmeans++~\cite{arthur2007kmeans++};
2) Entropy~\cite{entropy_al};
3) Marginal sampling~\cite{marginal_al};
4) Least Confidence (LC)~\cite{least_conf_al};
5) LfOSA~\cite{LfOSA};
6) OpenAL~\cite{OpenAL}.
Note that the first four methods are 
closed-set AL methods, while LfOSA and OpenAL are open-set AL methods.
All the existing methods used the same network as ours for a fair comparison.

Fig.~\ref{fig:comparision_crc100k_36}(a) shows the quantitative results on CRC100K, with labeled data increasing from 50 to 250 over five query rounds.
At the first query round, all methods used 50 randomly selected samples, achieving an average query precision of 33.60\% and model accuracy of 83.80\%.
In contrast, our proposed OpenPath achieved a query precision of 78.00\%, a significant improvement 
of 44.4 percentage points, and an accuracy of 95.26\%, an increase of 11.46 percentage points, demonstrating that our proposed warm-up strategy can select informative ID samples in the first query to train the model effectively.
In round 2, OpenPath further improved query precision by 20.40 percentage 
points, reaching 98.40\%. In comparison, the best existing method, 
LfOSA~\cite{LfOSA}, achieved a query precision of 93.20\%, but with a model 
accuracy of only 87.84\%. 
After 5 query rounds, 
the entropy-based method~\cite{entropy_al} performed the worst in terms of 
both query precision and model accuracy. 
This is because directly selecting samples with the highest entropy often 
results in choosing OOD samples, which do not contribute effectively to model 
training.
Our OpenPath achieved a model accuracy of 97.70\% with labeling ratio of 
0.25\%, close to the upper bound of 98.40\% achieved by fully 
supervised learning with 100\% labeled ID data.

Fig.~\ref{fig:comparision_crc100k_36}(b) shows the evaluation on the 
SkinTissue dataset, where the labeled pool grows from 200 to 1000 samples over 
5 query rounds.
OpenPath consistently achieved the highest model accuracy and accumulated 
query recall in each query round, with a final model accuracy of 97.68\%. It 
outperformed all the existing AL methods and was only 0.85 percentage points lower than fully supervised learning with 100\% labeled ID samples.


\subsection{Ablation Study}
\subsubsection{Effectiveness of warm-up strategy in the first round}

We conducted an ablation study on the two datasets 
to evaluate the contributions of VIDS and clustering in selecting representative ID samples in the first query.
For the CRC100K dataset, as shown in Table~\ref{tab:ablation1}, using VIDS in the first round significantly improved query precision and model accuracy, from 33.60\% to 78.40\% and from 83.80\% to 93.62\% compared with random selection, respectively. Incorporating clustering-based selection of representative samples from ID candidates further enhanced model accuracy to 95.26\%. 
Similar improvements were observed on the SkinTissue dataset, with VIDS enhancing both query precision and model accuracy, and clustering providing additional gains.

\subsubsection{Effectiveness of components in the subsequent rounds}


\begin{table}[t]
    \begin{minipage}{0.57\textwidth}
        \caption{Ablation study of warm-up strategy and comparison with existing cold start methods in the first query round. MAcc: Model Accuracy. QP: Query Precision.}
        \resizebox{0.97\textwidth}{!}{
        \centering
        \begin{tabular}{cc|cc|cc}
        \hline
        \multirow{2}{*}{VIDS} & \multirow{2}{*}{Cluster} & \multicolumn{2}{c|}{CRC100K} & \multicolumn{2}{c}{SkinTissue} \\ \cline{3-6}
         &     & MAcc (\%) & QP (\%) & MAcc (\%)  & QP (\%) \\ \hline
         &  & 83.80 & 33.60 & 90.51 & 17.30 \\
        \checkmark &  & 93.62 & \textbf{78.40} & 93.75 & 57.50 \\
         & \checkmark & 89.87 & 32.00 & 92.11 & 18.00 \\
        \checkmark & \checkmark & \textbf{95.26} & 78.00 & \textbf{94.28} & \textbf{58.50}\\ \hline
        \end{tabular}}
        \label{tab:ablation1}
    \end{minipage}
    \begin{minipage}{0.37\textwidth}
    \centering
    \begin{figure}[H]
        \includegraphics[width=\linewidth]{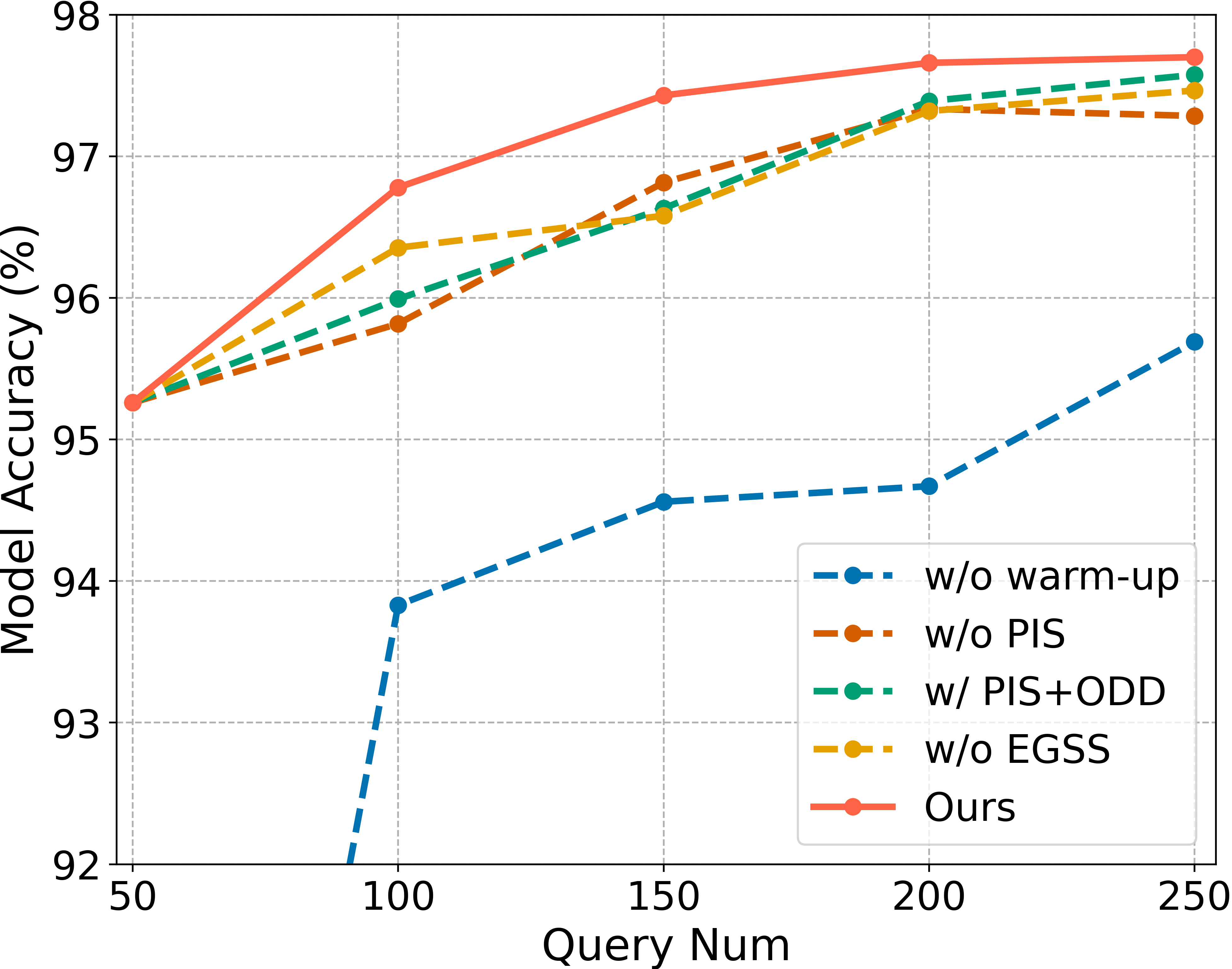}
        \caption{{Ablation study for all proposed components on the 
        CRC100K dataset.}}
        \label{fig:ablation_crc100k_36}
        \end{figure}
    \end{minipage}%
\end{table}

Fig.~\ref{fig:ablation_crc100k_36} shows ablation study results of all components, with the following notations: 
w/o warm-up indicates that 
random selection is used in the first round and w/ PIS + OOD refers to incorporating the centroid of OOD samples for calculating $d_i$.

The results showed that without the warm-up phase, 
the model experienced the largest decline in 
model accuracy, reaching a final model accuracy of 95.69\%, which is a drop of 2.01 percentage points. 
After just three query rounds, the proposed method significantly outperformed all other variants, which required more query rounds to reach that level of performance.

\section{Conclusion}
In this paper, we propose a novel open-set active learning method named OpenPath for pathology image classification. It effectively tackles two key challenges: 
selecting informative ID samples and addressing the cold start problem when there is no labeled data. 
Traditional AL methods often overlook the
OOD data, leading to inefficient annotation processes.
OpenPath leverages a pre-trained VLM with 
task-specific ID and OOD class prompts to mine ID samples, which addresses the cold start issue.
We also introduce a two-stage selection strategy to ensure that the query is both pure and informative. 
Comprehensive experiments showed that OpenPath can significantly 
enhance the purity of selected samples 
and outperform existing open-set AL methods.
In the future, it is of interest to apply our method to other open AL datasets. 

\section{Acknowledgment}
This work was supported in part by the National Natural Science
Foundation of China  under grant 62271115, and in part by the
Natural Science Foundation of Sichuan Province under grant 2025ZNSFSC0455.


\bibliographystyle{splncs04}
\bibliography{mybib}

\end{document}